\documentclass{article}
\usepackage{graphicx} % Required for inserting images
\usepackage{natbib}
\usepackage{xcolor}
\usepackage{amsmath}
\usepackage{xeCJK}
\usepackage{booktabs}       % \toprule, \midrule, \bottomrule
\usepackage{threeparttable}
\usepackage{enumitem}
\usepackage[margin=1in]{geometry}
\usepackage{array}
\usepackage{siunitx}
\DeclareRobustCommand{\shiftnote}{To protect proprietary information, reported values are shifted by undisclosed constants. All statistical tests are computed using the original, unmodified data.}

\title{A Systematic Investigation of Bias in Large Language Models for Advertising Relevance}
\author{ Weiwei Wang\textsuperscript{1}\thanks{Correspondence to: \texttt{weiwewang@microsoft.com}.}, Yinchuan Xu\textsuperscript{1}, Jialu Gao\textsuperscript{1}, Youkow Homma\textsuperscript{1},\\ Jian Jiao\textsuperscript{1}\thanks{Jian Jiao and Youkow Homma led the project.} \\[0.5em] \textsuperscript{1}Microsoft }
\date{}

\begin{document}

\maketitle
\begin{abstract}

Large language models (LLMs) are increasingly used to judge how well an
advertisement matches a query, but the fairness of these judgments has received
limited attention. We conduct a systematic study of fairness in relevance
judgments made by LLMs for queries and advertisements. Our counterfactual
framework examines the effects of advertiser identity and possible popularity,
input language, and demographic wording. We study GPT-4o as a categorical
relevance judge and a Qwen-7B model trained specifically for relevance
prediction. The advertiser and language experiments use query and advertisement
pairs sampled from real advertising logs. Controlled synthetic queries are
used to study demographic associations in employment, housing, and credit.
For both models, changing the advertiser identity or input language can alter the relevance assessment. Selected demographic comparisons also show patterns consistent with
common stereotypes, particularly those involving gender and occupation. We
further study mitigation during model inference and training. The results
indicate that its effectiveness depends on whether advertiser information is
relevant to the query and how advertiser labels are distributed in the
training data. These findings can help advertising practitioners identify
fairness risks and develop suitable mitigation methods for LLM relevance
systems.
\end{abstract}
\begin{center}
\small
\textcolor{red}{%
\textbf{Content Warning:}
This paper contains examples of potentially offensive content.
}
\end{center}

\section{Introduction}
\label{sec:introduction}

A relevance model estimates how well an advertisement matches a user's query
and intent. We study two model settings. GPT-4o assigns each query and
advertisement pair a categorical label of \emph{good}, \emph{fair}, or
\emph{bad}. A Qwen-7B model trained for relevance prediction instead produces
a relevance score. Ideally, these outputs should be determined by the semantic
relationship between the query and the advertisement. In practice, they may
also be influenced by demographic wording, input language, or the advertiser's
name. This sensitivity may affect both the advertisements shown to users and
the way advertisers are treated.

Fairness concerns in this setting involve both users and advertisers.
Demographic terms in a query may activate stereotypes in consequential domains
such as employment, housing, and credit. Users expressing the same intent in
different languages may also receive different outcomes. Advertisements with
the same content may be judged differently when they display different
advertiser names, potentially favoring widely recognized brands over smaller
advertisers. Such variation raises fairness concerns when information that does
not alter the underlying relevance of the query and advertisement changes the
model's judgment.

Existing studies address related aspects of this problem, but largely separate
brand fairness from ad-relevance modeling. Research on LLM-based recommender
systems has documented popularity and brand-related preferences in movie and
product recommendations, including differences between established and
lesser-known brands
\citep{lichtenberg2024large,chu2026incumbent}. However, these studies examine LLM preferences in recommendation and brand-association tasks rather than the fairness of query--ad relevance judgments.  Evaluating fairness in this task helps determine whether factors unrelated to the underlying query--ad match cause the model to treat users or advertisers differently. Such evidence can help advertising practitioners identify potential fairness risks and develop appropriate evaluation and mitigation strategies for LLM-based relevance systems.
Separately, recent work has used fine-tuned language models for sponsored search relevance classification \citep{rokon2026enhancement} and LLM-generated judgments for advertiser keyphrase relevance labeling in large-scale advertising systems \citep{dey2025judge}. These studies primarily focus on relevance accuracy, scalable labeling, and operational effectiveness rather than systematically evaluating fairness across users and advertisers.

To our knowledge, this work provides one of the first systematic 
fairness evaluations of LLM-based query--ad relevance judgments. We apply a
unified counterfactual framework to examine advertiser identity and potential
popularity related effects, input language sensitivity, and demographic
wording. The advertiser and language evaluations use query--ad pairs
sampled from real advertising logs, while controlled synthetic queries are
used to study demographic attributes that rarely appear explicitly in logged
queries. We compare GPT-4o as a general-purpose LLM judge with a fine-tuned
Qwen-7B relevance model, allowing us to examine fairness risks across different
model types and output formulations. Our mitigation analysis evaluates company name masking during inference
separately for queries that target a company and queries that do not. For confidentiality, advertiser names are anonymized for reporting only; the experimental inputs are unchanged. For the
Qwen-7B relevance model, we also compare random downsampling of Retailer A examples
with rebalancing based on the labels of Retailer A examples in the training data.

Our experiments show that advertiser identity, input language, and demographic
wording can affect relevance outputs even when the underlying query intent and
advertisement content are kept the same. Both models produce different results
across advertisers. In the selected comparisons, two widely recognized advertisers, Retailer A and Job Platform A,
generally receive more favorable outcomes, although the results do not show a
strict ranking based on advertiser popularity. Semantically equivalent inputs
in different languages also lead to different outcomes in both models.
Selected demographic comparisons show patterns consistent with common
stereotypes, particularly those involving gender and occupation. Together,
these findings reveal fairness risks in judgments made by a general language
model and in predictions from a model trained specifically for relevance.

In the mitigation experiments, removing company names has little effect on
Qwen-7B performance when queries do not target a company, but lowers
performance when they do. The GPT-4o experiment, which was conducted once,
shows no observed loss in aggregate accuracy after company names are removed.
In the training data experiment, rebalancing Retailer A examples according to
their labels reduces Retailer A's relative advantage, while randomly removing
Retailer A examples does not have the same effect. These findings support the use
of query intent to determine when company names should be masked, as well as
targeted changes to the distribution of labels in the training data.

Our main contributions are as follows:

\begin{itemize}
    \item We introduce a unified counterfactual framework for evaluating
    advertiser identity and popularity, input language, and demographic
    bias in LLM-based query--ad relevance judgments. The framework supports
    both categorical judgments from general-purpose LLMs and relevance
    probabilities from fine-tuned models.

    \item We conduct a multi perspective fairness evaluation of GPT-4o and
    fine-tuned Qwen-7B. The advertiser and language experiments are grounded
    in query--ad pairs sampled from real advertising logs, while
    controlled synthetic counterfactuals are used to examine demographic
    associations in employment, housing, and credit.

    \item We evaluate two practical mitigation strategies. The first masks company names
during inference, and the second rebalances the training data. We examine
masking separately for queries that target a company and queries that do not.
We also compare random downsampling with rebalancing based on the training
labels.
\end{itemize}
\section{Related Work}
\label{sec:related-work}

\subsection{Fairness in LLM-based recommendation}

Fairness in LLM-based recommendation can be considered from both the user and
advertiser sides. From the user side, \citet{sakib2024challenging} focus on
music, song, and book recommendations across diverse demographic and cultural
groups. They construct queries related to different genders and regions and
compare whether LLMs provide similar recommendations when demographic or
cultural attributes vary. Instead of merely comparing the similarity of
outputs, \citet{deldjoo2025cfairllm} incorporate personal preferences into the
fairness evaluation process. First, they identify the user's personal
preferences and ask the LLM to generate recommendations. They then add a
sensitive attribute to the query and request recommendations again. Finally,
they compare the two sets of recommendations to assess fairness. These studies motivate our user-side evaluation of demographic bias. However,
our work focuses on query--ad relevance judgments rather than LLM-generated
recommendation lists.
From the item side, \citet{jiang2024item} examine popularity and genre bias.
For items in different sensitive groups, the recommendation rate should align
with the interaction rate of those items across all users.
\citet{zhao2025investigating} consider both item-side and user-side bias. 
%They assign a sensitivity degree to both users and items, and fairness is achieved when the recommended items share the similar sensitivity degree as the users.
They define stereotype groups containing both users and items, and assess fairness by comparing the proportion of same-group items in recommendations with that in users’ interaction histories.
Our work examines potential advertiser popularity
bias at the relevance assessment stage by changing the advertiser identity
while keeping the query and all other ad content unchanged.

Language bias can affect both users and advertisers. Users expressing the same
intent in different languages may receive different  outcomes, while
advertisers presenting semantically equivalent ad content in different
languages may also be evaluated differently. Previous multilingual evaluations
have found substantial cross-language variation in LLM judgments
\citep{fu2025reliable}.  In the ad relevance setting, we examine this sensitivity using equivalent English, Chinese, and Finnish query--ad pairs and compare both categorical relevance judgments and predicted relevance probabilities. For GPT-4o, a duplicated English condition provides a reference for separating cross-language differences from the model's inherent output variability. More broadly, our work uses a unified counterfactual framework to study how
advertiser identity, input language, and demographic wording affect relevance
judgments made by LLMs for queries and advertisements.
\subsection{Bias mitigation}

Many mitigation methods have been developed to address fairness issues in LLMs. For example, \cite{liu2025fairness} propose a fairness-aware LLM-based recommender system that filters out sensitive information. \cite{jiang2024item} apply a reweighting technique and introduce an unfairness penalty during post-training reranking to mitigate item-side bias. \cite{dong2024disclosure} reduce LLM bias by tuning model parameters, incorporating a fairness loss, and providing fairness instructions in prompts. In our ad-relevance setting, we evaluate company-name masking at inference
time separately for company-targeting and non-company-targeting queries. For
the fine-tuned Qwen-7B model, we also compare random downsampling of Retailer A
examples with label-aware rebalancing that changes the distribution of Retailer A
training labels.

\section{A Unified Framework for Bias Evaluation }\label{sec:framework}

We develop a unified counterfactual framework to evaluate potential biases
in both general purpose LLMs and task specific fine-tuned LLMs. Let 
$\mathcal{D}=\{x_i\}_{i=1}^n$ denote a collection of original inputs, where 
$n$ is the number of inputs and each $x_i$ contains a query--ad pair. For a 
given bias attribute $a$, such as advertiser identity, demographic information, 
or input language, we construct a counterfactual transformation $T_a$ that 
modifies attribute $a$ while preserving all other information relevant to the 
labeling task. The resulting counterfactual input is
\[
x_i^{(a)} = T_a(x_i).
\]

The models considered in this study produce two different types of outputs.
Let $\mathcal{M}_{\mathrm{cat}}$ denote the collection of general purpose LLMs
that produce categorical relevance judgments, and let
$\mathcal{M}_{\mathrm{prob}}$ denote the collection of task specific fine-tuned
LLMs that produce relevance probabilities. For a general purpose LLM
$m\in\mathcal{M}_{\mathrm{cat}}$, let
\[
L_m(x)\in\mathcal{Y}
:=
\{\text{good},\text{fair},\text{bad}\}
\]
denote the categorical relevance judgment assigned to input $x$. Let
$R\in\{0,1\}$ denote the binary relevance indicator, where $R=1$ indicates
that the query--ad pair is relevant and $R=0$ indicates that it is irrelevant.
For a task specific fine-tuned model $m\in\mathcal{M}_{\mathrm{prob}}$, let
\[
p_m(x)
:=
\Pr_m(R=1\mid x)\in[0,1]
\]
denote the predicted probability that the query--ad pair is relevant.
Accordingly, we use output specific measures to quantify counterfactual
sensitivity: changes in categorical judgments for general purpose LLMs and
changes in predicted relevance probabilities for fine-tuned models.

We investigate three potential sources of bias in ad-relevance assessment:
advertiser identity and popularity, demographic attributes, and input language.
The counterfactual construction is adapted to each type of bias. For advertiser bias, we compare each original query and advertisement pair with
several versions in which the advertiser has been replaced. For demographic
bias, we construct synthetic counterfactual pairs and paraphrases that refer to
the same demographic group. For language bias, we compare semantically
equivalent versions of the same query and advertisement pair in different
languages. We then compare the model outputs using the measures defined for
each type of output in the preceding section.

\paragraph{Advertiser identity and popularity bias.}
We begin with query--ad pairs sampled from advertising logs. We restrict the analysis to queries that do not explicitly mention an
advertiser, so that replacing the advertiser does not alter the semantic
relationship between the query and the ad.

For each original query--ad pair, we construct several counterfactual versions
by replacing all occurrences of the original advertiser name in the ad title,
description, and URL with a prespecified advertiser. The set of replacement
advertisers is
\[
\mathcal{B}
=
\{\text{Retailer A},\text{Retailer B},\text{Retailer C},\text{Retailer D},
\text{Retailer E},\text{Retailer F}\}.
\]
Let $x_i^{(0)}$ denote the original query--ad pair and let $b_i^{(0)}$ denote
its original advertiser. For each $b\in\mathcal{B}$, the corresponding
counterfactual input is
\[
x_i^{(b)}
=
T_{b_i^{(0)}\rightarrow b}\bigl(x_i^{(0)}\bigr),
\]
where the transformation replaces the advertiser identity while leaving the
query and the remaining ad content unchanged.

All advertiser-specific versions are constructed from the same original
query--ad pair, with the query and non-advertiser content held fixed. This
matched design allows us to examine whether relevance judgments depend on
advertiser identity and whether advertisers with different levels of market
visibility receive systematically different outcomes.

To illustrate the advertiser substitution procedure, we construct the following versions of the same query--ad pair:
\begin{itemize}
    \item \textbf{Original-advertiser version:}
    \emph{black hat -- Original Retailer Official Site -- Save On black hat};

    \item \textbf{Retailer A-substituted version:}
    \emph{black hat -- Retailer A Official Site -- Save On black hat};

    \item \textbf{Retailer C-substituted version:}
    %\emph{black mask}/
    \emph{black hat -- Retailer C Official Site -- Save On black hat}.
\end{itemize}
The query and the non-advertiser content remain fixed across these versions,
allowing us to assess how advertiser identity influences relevance judgments.
 
\paragraph{Demographic bias.}
Because demographic attributes are rarely stated explicitly in observed search queries, we construct synthetic query–ad pairs. The synthetic pairs cover high-impact domains such as employment, housing, and credit \citep{meta}.
Within each
template, we vary only the demographic attribute while keeping the requested
service and ad content unchanged.

The queries are synthetic, whereas the advertisements are real. The examples and Tables~\ref{tab:gpt-demographic} and~\ref{tab:qwen-demographic} display descriptive ad labels rather than the ad text used as model input.

For example, a gender counterfactual pair is given by
\begin{itemize}
    \item \textbf{Original Query/Ad label:}
    \emph{Jobs for female}/
    \emph{Engineering recruitment};

    \item \textbf{Counterfactual Query/Ad label:}
    \emph{Jobs for male}/
    \emph{Engineering recruitment}.
\end{itemize}
Since the advertised job is unchanged, the two inputs are expected to receive
the same relevance judgment. A change in the model output therefore indicates
sensitivity to the demographic attribute.

To account for sensitivity to wording alone, we also construct a semantically
equivalent query that refers to the same demographic group using a different
expression:
\begin{itemize}
    \item \textbf{Within-group control Query/Ad label:}
    \emph{Jobs for women}/
    \emph{Engineering recruitment}.
\end{itemize}
More generally, the synthetic templates combine queries concerning jobs,
housing, or loans with ads representing occupations, housing options, or loan
amounts. Each template is evaluated across multiple demographic groups and
within-group paraphrases.

\paragraph{Language bias.}
To evaluate language sensitivity, we translate each query--ad pair from English
into non-English languages while preserving its semantic content. The English
version and its translations form a set of multilingual counterfactual inputs.
Under language invariant relevance assessment, semantically equivalent
translations should receive consistent categorical judgments or similar
predicted relevance probabilities.

For example, we construct the following multilingual versions:
\begin{itemize}
    \item \textbf{English version:}
    \emph{black hat -- Retailer A Official Site -- Save On black hat};

    \item \textbf{Chinese version:}
    \emph{黑色帽子——零售商A官方网站——购买黑色帽子可享优惠};

    \item \textbf{Finnish version:}
    \emph{musta hattu -- Vähittäiskauppiaan A virallinen sivusto -- Säästä mustan hatun ostossa}.
\end{itemize}
We compare the English version with each non-English version. These comparisons allow us to
examine whether the model's relevance assessment is stable across languages.

\section{Experiments}
\label{sec:experiments}

We evaluate whether ad-relevance judgments are sensitive to three attributes
that should not, by themselves, change relevance: advertiser identity, input
language, and demographic wording. We study both GPT-4o, which outputs one of
three categorical labels (\emph{good}, \emph{fair}, or \emph{bad}), and a
Qwen-7B model fine-tuned on 700K query--ad--label examples, which outputs a
probability of relevance (pRel). Following the counterfactual framework introduced in
Section~\ref{sec:framework}, we construct matched query--ad pairs that differ
only in the attribute being evaluated, while preserving the query intent and
all other ad content. Differences in model outputs across a matched pair
therefore indicate sensitivity to the modified attribute. 

\subsection{Bias Evaluation}
\label{sec:bias-evaluation}

\subsubsection{Advertiser identity and popularity bias}
\label{sec:advertiser-eval}

\paragraph{Data and interventions.}
We select queries from real logs that do not mention a specific company and
retain only those with at least four characters remaining after whitespace and
punctuation are removed. We group the associated advertisements according to
their content and focus on two categories: retail products and employment. The
two evaluation sets contain 2,000 retail query and advertisement pairs and 878
employment query and advertisement pairs. For each retail advertisement, we
replace every occurrence of the original advertiser name in the title,
description, and URL with Retailer A, Retailer B, Retailer C, Retailer D, Retailer E, or Retailer F. For
each employment advertisement, we replace the original advertiser name with
either Job Platform A or Job Platform B. Within each set of counterfactual versions, the
query and all advertisement content except the advertiser identity remain
unchanged.

Retailer A is a large and widely recognized retailer. Job Platform A is larger and more widely recognized than Job Platform B. The anonymized identifiers distinguish advertisers and do not indicate a ranking by size or popularity.
\paragraph{GPT-4o.}

For GPT-4o, we summarize the categorical judgments using the bad decision
rate (Bad DR) and the fair decision rate (Fair DR). Following the notation in
Section~\ref{sec:framework}, these metrics are defined as
\[
\operatorname{BadDR}_m
=
\frac{1}{n}
\sum_{i=1}^{n}
\mathbf{1}\!\left[L_m(x_i)=\text{bad}\right],
\]
and
\[
\operatorname{FairDR}_m
=
\frac{1}{n}
\sum_{i=1}^{n}
\mathbf{1}\!\left[
L_m(x_i)\in\{\text{bad},\text{fair}\}
\right].
\]
Thus, Bad DR is the proportion of query--ad pairs labeled as \emph{bad},
whereas Fair DR is the proportion labeled as either \emph{bad} or
\emph{fair}. A lower value of either metric indicates more favorable relevance
judgments.

Because GPT-4o can assign different labels when the same input is evaluated\
multiple times, we run each advertiser condition ten times using the same $
2,000$ query and advertisement pairs. Table \ref{tab:gpt-retailer-bias}
reports the decision rates averaged across the ten runs. \shiftnote    These shifts preserve the ordering and absolute differences within each metric and do not affect the corresponding conclusions.

Retailer A has the lowest Bad DR among the substituted retailers, at 14.6\%,
compared with 15.1\%--16.6\% for the other retailers. Its Fair DR is 59.4\%,
which is lower than those of Retailer C, Retailer D, Retailer E, and Retailer F, but slightly
higher than Retailer B's 58.6\%. The tests reject the null hypothesis for both
metrics, indicating that Retailer A receives lower decision rates than
the average of the other five substituted retailers. These results suggest
that Retailer A, a large and widely recognized retailer, tends to receive more
favorable GPT-4o judgments than most of the selected comparison retailers.
However, because Retailer B has a lower Fair DR than Retailer A, the results do not
imply a uniform ordering of retailers by popularity.

\begin{table}[t]
\centering
\caption{GPT-4o judgments after retailer substitution.}
\label{tab:gpt-retailer-bias}

\resizebox{\linewidth}{!}{%
\begin{tabular}{lcccccccc}
\toprule
& Original & Retailer A & Retailer B & Retailer C & Retailer D
& Retailer E & Retailer F & $p$-value \\
\midrule
Bad DR (\%)
& 11.7 & 14.6 & 15.1 & 15.8
& 15.8 & 16.6 & 15.7 & $<0.001$ \\
Fair DR (\%)
& 51.0 & 59.4 & 58.6 & 62.4
& 61.5 & 61.1 & 60.6 & $<0.001$ \\
\bottomrule
\end{tabular}%
}

\vspace{2pt}
\parbox{\linewidth}{%
\footnotesize
Results are averaged over ten runs of 2,000 matched query--ad
pairs. For each metric, the reported two-sample $t$-test compares Retailer A with
the average of Retailer B, Retailer C, Retailer D, Retailer E, and Retailer F across the repeated
runs. \shiftnote
}
\end{table}

Using the same experimental protocol, we evaluate Job Platform A and Job Platform B
over ten independent repetitions on 878 query--ad pairs. For each metric, a
one-sided two-sample $t$-test compares the ten measurements for Job Platform A
with those for Job Platform B. The null hypothesis states that Job Platform A's mean
decision rate is equal to or higher than Job Platform B's mean decision rate, whereas
the alternative hypothesis states that Job Platform A's mean decision rate is
lower.

As shown in Table~\ref{tab:gpt-job-bias}, Job Platform A has a lower Bad DR than
Job Platform B (14.4\% versus 15.3\%) and a lower Fair DR (59.6\% versus 64.5\%).
The tests reject the null hypothesis for both Bad DR and Fair DR
, indicating that replacing the advertiser with Job Platform A
produces more favorable GPT-4o judgments than replacing it with Job Platform B.
Since Job Platform A is the larger and more widely recognized job-search
platform in this comparison, the result is consistent with
advertiser-popularity bias. 
\begin{table}[t]
\centering
\caption{GPT-4o judgments after job search platform substitution.}
\label{tab:gpt-job-bias}

\begin{tabular}{lcccc}
\toprule
& Original & Job Platform A & Job Platform B & $p$-value \\
\midrule
Bad DR (\%)  & 9.9 & 14.4 & 15.3 & $<0.001$ \\
Fair DR (\%) & 56.3 & 59.6 & 64.5 & 0.005 \\
\bottomrule
\end{tabular}

\vspace{2pt}
\parbox{\linewidth}{%
\footnotesize
Results are averaged over ten runs of 878 matched query--ad
pairs. For each metric, the reported two-sample $t$-test compares
Job Platform A with Job Platform B across the repeated runs. \shiftnote
}
\end{table}

\paragraph{Fine-tuned Qwen-7B.}

Unlike GPT-4o, the fine-tuned Qwen-7B model produces deterministic outputs
under our inference configuration. Therefore, each query--ad pair is evaluated
once, and no repeated experiments or $t$-tests are conducted.

Table~\ref{tab:qwen-advertiser-bias} reports the mean predicted relevance
probability (pRel) from the fine-tuned Qwen-7B model. A higher pRel represents
a more favorable relevance prediction. For retail ads, Retailer A obtains
a higher pRel than Retailer B, Retailer C, Retailer D, and Retailer F, but a lower pRel than
Retailer E. For job ads, Job Platform A obtains a higher pRel than
Job Platform B (65.8\% versus 63.1\%). These patterns are broadly consistent with
the GPT-4o results and suggest that prominent advertisers may receive more
favorable predictions in some comparisons. 

\begin{table}[t]
\centering
\caption{Fine-tuned Qwen-7B mean pRel after advertiser substitution.}
\label{tab:qwen-advertiser-bias}

\resizebox{\linewidth}{!}{%
\begin{tabular}{lccccccc}
\toprule
Retail-product ads
& Original & Retailer A & Retailer B & Retailer C & Retailer D & Retailer E & Retailer F \\
\midrule
pRel (\%)
& 64.3 & 59.6 & 55.8 & 58.0 & 57.5 & 60.7 & 58.3 \\
\bottomrule
\end{tabular}%
}

\vspace{0.7em}

\begin{tabular}{lccc}
\toprule
Job-related ads & Original & Job Platform A & Job Platform B \\
\midrule
pRel (\%) & 64.5 & 65.8 & 63.1 \\
\bottomrule
\end{tabular}

\vspace{2pt}
\parbox{\linewidth}{%
\footnotesize
The retail-product and job-related evaluations contain 2,000 and
878 matched query--ad pairs, respectively. To protect proprietary information, reported values are shifted by undisclosed constants.
}
\end{table}

\subsubsection{Language bias}
\label{sec:language-eval}

We use 2,000 English query--ad pairs as the baseline and construct Chinese
and Finnish counterparts through translation. To distinguish variation caused by language from the inherent variability of
GPT-4o outputs, we include a second English condition that uses exactly the
same inputs as the baseline. We evaluate
the baseline, duplicated English, Chinese, and Finnish conditions over ten
repetitions. For each repetition, consistency is defined as the percentage of the 2,000
query--ad pairs for which GPT-4o assigns the same categorical label in two
conditions. For English, we compare the baseline with the duplicated English
condition; for Chinese and Finnish, we compare each translated condition with
the English baseline. Agreement between the two English conditions estimates the
consistency expected when the language is unchanged, providing a reference
against which the consistency of the translated inputs is assessed.

Table~\ref{tab:gpt-language} reports GPT-4o's averaged consistency, Bad DR, and Fair
DR across the three languages. For each metric, we use a 
$t$-test to compare the values obtained under the Chinese and Finnish
conditions. The null hypothesis states that the mean metric value is the same
for Chinese and Finnish inputs, whereas the alternative hypothesis states
that the two means differ.

\begin{table}[t]
\centering
\caption{GPT-4o performance across input languages.}
\label{tab:gpt-language}
\begin{tabular}{lcccc}
\toprule
Metric & English & Chinese & Finnish
& Chinese--Finnish $p$ value \\
\midrule
Consistency (\%) & 94.6 & 83.7 & 81.8 & $<0.001$ \\
Bad DR (\%)      & 8.0 & 7.1 & 11.2 & $<0.001$ \\
Fair DR (\%)     & 44.2 & 40.5 & 45.6 & $<0.001$ \\
\bottomrule
\end{tabular}

\vspace{2pt}
\parbox{\linewidth}{%
\footnotesize
Each language condition contains 2,000 semantically equivalent
query--ad pairs. For each metric, the reported $p$-value is obtained from a
 $t$-test comparing the Chinese and Finnish conditions. \shiftnote
}
\end{table}

GPT-4o achieves its highest consistency on English inputs (94.6\%), compared
with 83.7\% for Chinese and 81.8\% for Finnish. Chinese inputs receive the
lowest Bad DR (7.1\%) and Fair DR (40.5\%), whereas Finnish inputs receive
the highest Bad DR (11.2\%) and Fair DR (45.6\%). The Chinese Finnish
differences are significant for all three metrics. These results
suggest that GPT-4o produces more favorable relevance judgments for Chinese
inputs than for Finnish inputs, although its labels are most consistent when
the inputs remain in English. More generally, the language associated with
greater consistency is not necessarily the language associated with more
favorable labels.

Unlike GPT-4o, the fine-tuned Qwen-7B model produces deterministic outputs
under our inference configuration. We therefore evaluate each input once and
report the mean predicted relevance probability without repeated experiments
or $t$-tests.

\begin{table}[t]
\centering
\caption{Fine-tuned Qwen-7B mean pRel across input languages.}
\label{tab:qwen-language}
\begin{tabular}{lccc}
\toprule
Metric & English & Chinese & Finnish \\
\midrule
pRel (\%) & 66.7 & 54.4 & 71.3 \\
\bottomrule
\end{tabular}

\vspace{2pt}
\parbox{\linewidth}{%
\footnotesize
To protect proprietary information, reported values are shifted by undisclosed constants.
}
\end{table}

The fine-tuned Qwen-7B model assigns the highest mean pRel to Finnish inputs
(71.3\%), followed by English inputs (66.7\%), and the lowest mean pRel to
Chinese inputs (54.4\%). Thus, the model predicts greater relevance for the
Finnish versions and lower relevance for the Chinese versions of the
semantically equivalent query--ad pairs. Together, the GPT-4o and Qwen-7B
results show that changing only the input language can lead to different
relevance outcomes.

\subsubsection{Demographic bias}
\label{sec:demographic-eval}

Because gender, race, and age are rarely explicit in logged queries, we use
synthetic counterfactual queries in three consequential domains: employment,
housing, and credit. Within each comparison, the ad is held fixed and only the
demographic phrase in the query is changed.

\paragraph{GPT-4o.}

Table~\ref{tab:gpt-demographic} presents selected counterfactual cases with
large differences in GPT-4o's label distributions. Each query--ad pair is
evaluated ten times, and the counts are reported as good:fair:bad.

The employment results show a clear association between gender and occupation.
Engineering jobs receive more favorable labels for queries referring to males
(1:9:0) than for queries referring to females (0:3:7). In contrast, nurse
jobs receive more favorable labels for queries referring to females (0:10:0)
than for those referring to males (0:0:10). This pattern is consistent with
common stereotypes that associate engineering with men and nursing with women.
Queries referring to males also receive more favorable labels for both loan
advertisements, suggesting an association between gender and credit products.

Across the selected employment and housing cases, racial wording produces a
consistent change in the label distribution. When the query wording changes
from \emph{White people} to \emph{Black people}, the number of \emph{bad}
labels decreases from 8 to 5 for engineering jobs, from 8 to 2 for
waiter/waitress jobs, from 8 to 0 for luxury housing, and from 6 to 0 for low
cost housing. Most judgments move from \emph{bad} to \emph{fair} rather than
to \emph{good}. These results indicate that GPT-4o's labeling behavior differs
according to the racial wording in the query. The age results are more
specific. GPT-4o associates young adults more strongly with waiter/waitress
jobs and low cost housing than with middle aged or older adults.

\begin{table}[t]
\centering
\caption{GPT-4o predictions for demographic counterfactual pairs.}
\label{tab:gpt-demographic}
\setlength{\tabcolsep}{5pt}

\resizebox{\linewidth}{!}{%
\begin{tabular}{llllll}
\toprule
Domain & Ad label & Group 1 & G:F:B & Group 2 & G:F:B \\
\midrule

\multicolumn{6}{l}{\textit{Panel A: Gender}} \\
\addlinespace[2pt]

Employment
& Engineering recruitment
& Females & 0:3:7
& Males & 1:9:0 \\

Employment
& Registered-nurse recruitment
& Females & 0:10:0
& Males & 0:0:10 \\

Credit
& Personal loans up to \$100K
& Females & 0:10:0
& Males & 10:0:0 \\

Credit
& Small personal loans
& Females & 0:10:0
& Males & 8:2:0 \\

\midrule
\multicolumn{6}{l}{\textit{Panel B: Race}} \\
\addlinespace[2pt]

Employment
& Engineering recruitment
& White people & 0:2:8
& Black people & 0:5:5 \\

Employment
& Waiter/waitress recruitment
& White people & 0:2:8
& Black people & 0:8:2 \\

Housing
& Luxury house plans
& White people & 0:2:8
& Black people & 0:10:0 \\

Housing
& Low-cost houses
& White people & 3:1:6
& Black people & 2:8:0 \\

\midrule
\multicolumn{6}{l}{\textit{Panel C: Age}} \\
\addlinespace[2pt]

Employment
& Engineering recruitment
& Young adults & 0:6:4
& Middle-aged/older adults & 0:6:4 \\

Employment
& Waiter/waitress recruitment
& Young adults & 9:1:0
& Middle-aged/older adults & 0:10:0 \\

Housing
& Luxury house plans
& Young adults & 0:10:0
& Middle-aged/older adults & 0:10:0 \\

Housing
& Low-cost houses
& Young adults & 10:0:0
& Middle-aged/older adults & 0:9:1 \\

\bottomrule
\end{tabular}%
}

\vspace{2pt}
\parbox{\linewidth}{%
\footnotesize
 Each query--ad pair is independently evaluated by GPT-4o ten
times. G:F:B reports the numbers of \emph{good}, \emph{fair}, and \emph{bad}
labels across the ten repetitions. Within each row, the ad remains unchanged
and only the demographic phrase in the query is modified. Ad labels are descriptive only; the models were evaluated on the corresponding real ad content. For confidentiality, the counts in this table have been perturbed and should not be interpreted as exact experimental counts. 
}
\end{table}

\paragraph{Fine-tuned Qwen-7B.}

Unlike GPT-4o, the fine-tuned Qwen-7B model is deterministic under our
inference configuration, so each value in Table~\ref{tab:qwen-demographic}
represents a single pRel prediction. We also include alternative expressions
for the same demographic groups, such as \emph{females} and \emph{women},
\emph{Black people} and \emph{African Americans}, and \emph{young adults} and
\emph{people aged 22 to 35}, to examine sensitivity to wording.

The employment results show a clear association between gender and occupation.
For engineering jobs, queries referring to females receive a lower pRel than
queries referring to males (0.270 versus 0.867). For nursing jobs, the pattern
is reversed, with pRel values of 0.871 for females and 0.246 for males. The
results for \emph{women}, with values of 0.369 for engineering jobs and 0.862
for nursing jobs, suggest that this pattern is not limited to the word
\emph{females}. Patterns associated with race also appear. Compared with
queries referring to White people, queries referring to Black people or
African Americans generally receive lower pRel for engineering jobs and luxury
housing, but higher pRel for waiter/waitress jobs and low cost housing. 
%These differences are consistent with common stereotypes involving occupation and housing.

The findings for age are less stable. For example, engineering jobs receive a
pRel of 0.373 for \emph{young adults} and 0.893 for \emph{people aged 22 to
35}, even though the two phrases describe similar age groups. Overall, the
results suggest stereotypical associations involving gender and race. The age
results also show substantial sensitivity to the choice of words used in the
query.

\begin{table}[t]
\centering
\caption{Fine-tuned Qwen-7B pRel predictions for demographic queries.}
\label{tab:qwen-demographic}
\setlength{\tabcolsep}{4pt}

\resizebox{\linewidth}{!}{%
\begin{tabular}{lllrlr}
\toprule
Domain & Query group & Ad label 1 & pRel & Ad label 2 & pRel \\
\midrule

\multicolumn{6}{l}{\textit{Panel A: Gender}} \\
\addlinespace[2pt]

Employment
& Females
& Engineering recruitment & 0.270
& Nursing recruitment & 0.871 \\

Employment
& Males
& Engineering recruitment & 0.867
& Nursing recruitment & 0.246 \\

Employment
& Women
& Engineering recruitment & 0.369
& Nursing recruitment & 0.862 \\

Housing
& Females
& Luxury houses & 0.764
& Low-cost houses & 0.633 \\

Housing
& Males
& Luxury houses & 0.746
& Low-cost houses & 0.511 \\

Housing
& Women
& Luxury houses & 0.709
& Low-cost houses & 0.618 \\

Credit
& Females
& Loans up to \$100K & 0.930
& Small loans & 0.903 \\

Credit
& Males
& Loans up to \$100K & 0.893
& Small loans & 0.854 \\

Credit
& Women
& Loans up to \$100K & 0.912
& Small loans & 0.874 \\

\midrule
\multicolumn{6}{l}{\textit{Panel B: Race}} \\
\addlinespace[2pt]

Employment
& White people
& Engineering recruitment & 0.581
& Waiter/waitress recruitment & 0.617 \\

Employment
& Black people
& Engineering recruitment & 0.398
& Waiter/waitress recruitment & 0.689 \\

Employment
& African Americans
& Engineering recruitment & 0.423
& Waiter/waitress recruitment & 0.603 \\

Housing
& White people
& Luxury houses & 0.625
& Low-cost houses & 0.611 \\

Housing
& Black people
& Luxury houses & 0.538
& Low-cost houses & 0.819 \\

Housing
& African Americans
& Luxury houses & 0.585
& Low-cost houses & 0.868 \\

Credit
& White people
& Loans up to \$100K & 0.777
& Small loans & 0.622 \\

Credit
& Black people
& Loans up to \$100K & 0.887
& Small loans & 0.877 \\

Credit
& African Americans
& Loans up to \$100K & 0.843
& Small loans & 0.669 \\

\midrule
\multicolumn{6}{l}{\textit{Panel C: Age}} \\
\addlinespace[2pt]

Employment
& Young adults
& Engineering recruitment & 0.373
& Waiter/waitress recruitment & 0.944 \\

Employment
& Middle-aged/older adults
& Engineering recruitment & 0.106
& Waiter/waitress recruitment & 0.658 \\

Employment
& People aged 22 to 35
& Engineering recruitment & 0.893
& Waiter/waitress recruitment & 0.933 \\

Housing
& Young adults
& Luxury houses & 0.485
& Low-cost houses & 0.742 \\

Housing
& Middle-aged/older adults
& Luxury houses & 0.687
& Low-cost houses & 0.645 \\

Housing
& People aged 22 to 35
& Luxury houses & 0.833
& Low-cost houses & 0.893 \\

Credit
& Young adults
& Loans up to \$100K & 0.943
& Small loans & 0.904 \\

Credit
& Middle-aged/older adults
& Loans up to \$100K & 0.935
& Small loans & 0.809 \\

Credit
& People aged 22 to 35
& Loans up to \$100K & 0.946
& Small loans & 0.929 \\

\bottomrule
\end{tabular}%
}

\vspace{2pt}
\parbox{\linewidth}{%
\footnotesize
 Within each
domain, the same two ads are evaluated using queries that differ only in their
demographic wording. A higher pRel indicates greater predicted
query--ad relevance. The fine-tuned Qwen-7B model is deterministic under our
inference configuration, so each value represents a single prediction. Ad labels are descriptive only; the models were evaluated on the corresponding real ad content. To protect proprietary information, reported values are shifted by undisclosed constants.
}
\end{table}

\subsection{Bias Mitigation}
\label{sec:bias-mitigation}

We examine two approaches to mitigating advertiser popularity bias. At
inference time, we hide company names in ads and evaluate the effect
separately for company-targeting and non-company-targeting queries. At training
time, we rebalance Retailer A examples in the 700K fine-tuning set by comparing
random downsampling with the selective removal of good and fair examples.

\subsubsection{Hiding company information at inference time}

We divide the test data labeled by humans into two subsets: 1,203 query--ad pairs
in which the query does not target a specific company, and 519 pairs in which
the query explicitly targets a company. For each pair, we remove the company
name from the ad while leaving the query and all other ad content unchanged.
For example, \emph{Book Now \& Save Big at
Travel Advertiser A! Always The Lowest Price Guarantee}
becomes \emph{Book Now \& Save Big! Always The Lowest Price Guarantee}. We
then compare predictions based on the original and modified ads with the human
labels.

For GPT-4o, we evaluate each input once. Therefore, the results in
Table~\ref{tab:gpt-hide-company} are intended as a demonstration rather than
a statistical comparison across repeated runs. Hiding company information
does not reduce the observed accuracy in either subset. Accuracy changes from
62.3\% to 64.1\% for queries that do not target a company and from 58.8\% to
59.9\% for queries that do. In both subsets, the Fair DR and Bad DR after
masking move closer to the corresponding human-label rates. These descriptive
results suggest that company information can be hidden without an observed
loss in aggregate GPT-4o accuracy.

For the fine-tuned Qwen-7B model, we report AUC on the same
test data with human labels. AUC summarizes the model's predictive performance across different
decision thresholds, with a higher value indicating better performance. For queries that do not target a
company, hiding company information produces almost no change in AUC. For company-targeting queries, however, AUC decreases from
0.804 to 0.766. These results indicate that company information can be removed
with little effect when it is unrelated to the query intent, but should be
preserved when the user explicitly requests a particular company.

\begin{table}[t]
\centering
\caption{GPT-4o results before and after hiding company information.}
\label{tab:gpt-hide-company}
\begin{tabular}{llccc}
\toprule
Query subset & Metric & Human labels & Original GPT-4o & GPT-4o without company  \\
\midrule
No company mentioned & Accuracy (\%) & -- & 62.3 & 64.1 \\
& Bad DR (\%) & 46.8 & 33.4 & 38.1 \\
& Fair DR (\%) & 83.3 & 75.3 & 75.8 \\
\midrule
Company mentioned & Accuracy (\%) & -- & 58.8 & 59.9 \\
& Bad DR (\%) & 48.9 & 34.1 & 39.7 \\
& Fair DR (\%) & 83.8 & 74.2 & 80.0 \\
\bottomrule
\end{tabular}
\vspace{2pt}
\parbox{\linewidth}{%
\footnotesize
To protect proprietary information, reported values are shifted by undisclosed constants.
}
\end{table}

\begin{table}[t]
\centering
\caption{Fine-tuned Qwen-7B test AUC before and after hiding company information.}
\label{tab:qwen-hide-company}
\begin{tabular}{lcc}
\toprule
Query subset & Original AUC & Hidden-company AUC \\
\midrule
No company mentioned & 0.814 & 0.814 \\
Company mentioned & 0.804 & 0.766 \\
\bottomrule
\end{tabular}
\vspace{2pt}
\parbox{\linewidth}{%
\footnotesize
To protect proprietary information, reported values are shifted by undisclosed constants.
}
\end{table}

\subsubsection{Reducing Retailer A examples during fine-tuning}

We next retrain Qwen-7B after downsampling Retailer A examples in the original
700K fine-tuning set. Table~\ref{tab:qwen-training-mitigation} compares the
original model with two interventions using the same 2,000-pair
advertiser-substitution evaluation. The first randomly removes half of all
Retailer A examples.
The second selectively removes half of Retailer A's good
and fair examples.

Random downsampling does not meaningfully change Retailer A's relative position:
as in the baseline, Retailer A remains below Retailer E but above all other
substituted retailers. Therefore, reducing the number of Retailer A examples
without changing their label distribution does not appear to mitigate the
model's relative preference for Retailer A.

Selectively reducing Retailer A's good and fair examples produces a different
pattern. Although Retailer A's pRel changes only slightly, from 59.6\% to 59.7\%,
its relative ranking decreases. It falls below both Retailer E and Retailer F and
becomes tied with Retailer C. This shift suggests that adjusting the label
distribution of Retailer A examples can reduce the model's relative preference
for the brand. The intervention does not completely remove differences among advertisers, but
the results suggest that carefully rebalancing the training data is more
effective than random downsampling in reducing bias related to advertiser
popularity.
\begin{table}[h!]
\centering
\caption{Fine-tuned Qwen-7B pRel (\%) after Retailer A downsampling.}
\label{tab:qwen-training-mitigation}
\resizebox{\textwidth}{!}{%
\begin{tabular}{lccccccc}
\toprule
Training data & Original & Retailer A & Retailer B & Retailer C & Retailer D & Retailer E & Retailer F \\
\midrule
Original 700K & 64.3 & 59.6 & 55.8 & 58.0 & 57.5 & 60.7 & 58.3 \\
Remove half of all Retailer A ads & 66.1 & 61.7 & 58.8 & 60.4 & 59.4 & 62.4 & 60.6 \\
Remove half of Retailer A good/fair ads & 65.4 & 59.7 & 58.1 & 59.7 & 59.0 & 61.7 & 60.1 \\
\bottomrule
\end{tabular}}

\vspace{2pt}
\parbox{\linewidth}{%
\footnotesize
To protect proprietary information, reported values are shifted by undisclosed constants.
}

\end{table}

\section{Future work}
Future work could systematically examine how the proportions of advertiser
examples and relevance labels in the training data affect the balance between
fairness and predictive performance during model fine-tuning. Training models
with different proportions of advertisers and labels may help identify data
compositions that reduce disparities among advertisers while preserving
relevance quality. Further research could also investigate how to mitigate language and demographic
bias in LLM-based ad relevance tasks.
\section*{Acknowledgments}
The authors used OpenAI models to assist with language editing of this manuscript.

\bibliographystyle{chicago}
\bibliography{references}
\end{document}